\documentclass[10pt,twocolumn,letterpaper]{article}

\usepackage[pagenumbers]{iccv} 

\usepackage{hyperref}
\usepackage{pifont}
\usepackage{multirow}
\usepackage{graphicx}
\usepackage{wrapfig}
\usepackage{arydshln}
\usepackage{textcomp}
\usepackage[dvipsnames,table]{xcolor}

\usepackage{float}

\definecolor{White}{rgb}{1.,0.,1.}
\definecolor{first}{rgb}{.8,.0,.0}
\definecolor{second}{rgb}{.0,.6,.0}
\definecolor{third}{rgb}{.0,.0,.8}

\definecolor{ceiling}{RGB}{214,  38, 40}
\definecolor{floor}{RGB}{43, 160, 4}
\definecolor{wall}{RGB}{158, 216, 229}
\definecolor{window}{RGB}{114, 158, 206}
\definecolor{chair}{RGB}{204, 204, 91}
\definecolor{bed}{RGB}{255, 186, 119}
\definecolor{sofa}{RGB}{147, 102, 188}
\definecolor{table}{RGB}{30, 119, 181}
\definecolor{tvs}{RGB}{160, 188, 33}
\definecolor{furniture}{RGB}{255, 127, 12}
\definecolor{objects}{RGB}{196, 175, 214}

\definecolor{car}{rgb}{0.39215686, 0.58823529, 0.96078431}
\definecolor{bicycle}{rgb}{0.39215686, 0.90196078, 0.96078431}
\definecolor{motorcycle}{rgb}{0.11764706, 0.23529412, 0.58823529}
\definecolor{truck}{rgb}{0.31372549, 0.11764706, 0.70588235}
\definecolor{othervehicle}{rgb}{0.39215686, 0.31372549, 0.98039216}
\definecolor{person}{rgb}{1.        , 0.11764706, 0.11764706}
\definecolor{bicyclist}{rgb}{1.        , 0.15686275, 0.78431373}
\definecolor{motorcyclist}{rgb}{0.58823529, 0.11764706, 0.35294118}
\definecolor{road}{rgb}{1.        , 0.        , 1.        }
\definecolor{parking}{rgb}{1.        , 0.58823529, 1.        }
\definecolor{sidewalk}{rgb}{0.29411765, 0.        , 0.29411765}
\definecolor{otherground}{rgb}{0.68627451, 0.        , 0.29411765}
\definecolor{building}{rgb}{1.        , 0.78431373, 0.        }
\definecolor{fence}{rgb}{1.        , 0.47058824, 0.19607843}
\definecolor{vegetation}{rgb}{0.        , 0.68627451, 0.        }
\definecolor{trunk}{rgb}{0.52941176, 0.23529412, 0.        }
\definecolor{terrain}{rgb}{0.58823529, 0.94117647, 0.31372549}
\definecolor{pole}{rgb}{1.        , 0.94117647, 0.58823529}
\definecolor{trafficsign}{rgb}{1.        , 0.        , 0.        }
\definecolor{otherstructure}{rgb}{0.98039215, 0.58823529, 0.}
\definecolor{otherobject}{rgb}{0.19607843, 1.        , 1.        }

\makeatletter
\newcommand{\car@semkitfreq}{3.92}
\newcommand{\bicycle@semkitfreq}{0.03}
\newcommand{\motorcycle@semkitfreq}{0.03}
\newcommand{\truck@semkitfreq}{0.16}
\newcommand{\othervehicle@semkitfreq}{0.20}
\newcommand{\person@semkitfreq}{0.07}
\newcommand{\bicyclist@semkitfreq}{0.07}
\newcommand{\motorcyclist@semkitfreq}{0.05}
\newcommand{\road@semkitfreq}{15.30}
\newcommand{\parking@semkitfreq}{1.12}
\newcommand{\sidewalk@semkitfreq}{11.13}
\newcommand{\otherground@semkitfreq}{0.56}
\newcommand{\building@semkitfreq}{14.1}
\newcommand{\fence@semkitfreq}{3.90}
\newcommand{\vegetation@semkitfreq}{39.3}
\newcommand{\trunk@semkitfreq}{0.51}
\newcommand{\terrain@semkitfreq}{9.17}
\newcommand{\pole@semkitfreq}{0.29}
\newcommand{\trafficsign@semkitfreq}{0.08}
\newcommand{\semkitfreq}[1]{{\csname #1@semkitfreq\endcsname}}

\newcommand{\car@sscbkitfreq}{2.85}
\newcommand{\bicycle@sscbkitfreq}{0.01}
\newcommand{\motorcycle@sscbkitfreq}{0.01}
\newcommand{\truck@sscbkitfreq}{0.16}
\newcommand{\othervehicle@sscbkitfreq}{5.75}
\newcommand{\person@sscbkitfreq}{0.02}
\newcommand{\road@sscbkitfreq}{14.98}
\newcommand{\parking@sscbkitfreq}{2.31}
\newcommand{\sidewalk@sscbkitfreq}{6.43}
\newcommand{\otherground@sscbkitfreq}{2.05}
\newcommand{\building@sscbkitfreq}{15.67}
\newcommand{\fence@sscbkitfreq}{0.96}
\newcommand{\vegetation@sscbkitfreq}{41.99}
\newcommand{\terrain@sscbkitfreq}{7.10}
\newcommand{\pole@sscbkitfreq}{0.22}
\newcommand{\trafficsign@sscbkitfreq}{0.06}
\newcommand{\otherstructure@sscbkitfreq}{4.33}
\newcommand{\otherobject@sscbkitfreq}{0.28}
\newcommand{\sscbkitfreq}[1]{{\csname #1@sscbkitfreq\endcsname}}

\definecolor{iccvblue}{rgb}{0.21,0.49,0.74}

\def\paperID{535} 
\def\confName{ICCV}
\def\confYear{2025}

\title{Disentangling Instance and Scene Contexts for 3D Semantic Scene Completion}

\author{
    Enyu Liu$^{}$\thanks{Equal contribution.} \quad
    En Yu$^{*}$ \quad Sijia Chen \quad  Wenbing Tao $^{}$\thanks{Corresponding author.} \\
    Huazhong University of Science and Technology \\
    {\tt\small \{eyliu,yuen,sijiachen,wenbingtao\}@hust.edu.cn}
}

\begin{document}
\maketitle
\begin{abstract}
3D Semantic Scene Completion (SSC) has gained increasing attention due to its pivotal role in 3D perception. Recent advancements have primarily focused on refining voxel-level features to construct 3D scenes. However, treating voxels as the basic interaction units inherently limits the utilization of class-level information, which is proven critical for enhancing the granularity of completion results. To address this, we propose \textbf{D}isentangling \textbf{I}nstance and \textbf{S}cene \textbf{C}ontexts (\textbf{DISC}), a novel dual-stream paradigm that enhances learning for both instance and scene categories through separated optimization. Specifically, we replace voxel queries with discriminative class queries, which incorporate class-specific geometric and semantic priors. Additionally, we exploit the intrinsic properties of classes to design specialized decoding modules, facilitating targeted interactions and efficient class-level information flow. Experimental results demonstrate that DISC achieves state-of-the-art (SOTA) performance on both SemanticKITTI and SSCBench-KITTI-360 benchmarks, with mIoU scores of 17.35 and 20.55, respectively. Remarkably, DISC even outperforms multi-frame SOTA methods using only single-frame input and significantly improves instance category performance, surpassing both single-frame and multi-frame SOTA instance mIoU by \textbf{17.9\%} and \textbf{11.9\%}, respectively, on the SemanticKITTI hidden test.
The code is available at \url{https://github.com/Enyu-Liu/DISC}.
\end{abstract}
\section{Introduction}
\label{sec:intro}

\begin{figure}[!t]
    \centering
    \includegraphics[width=1.0\linewidth]{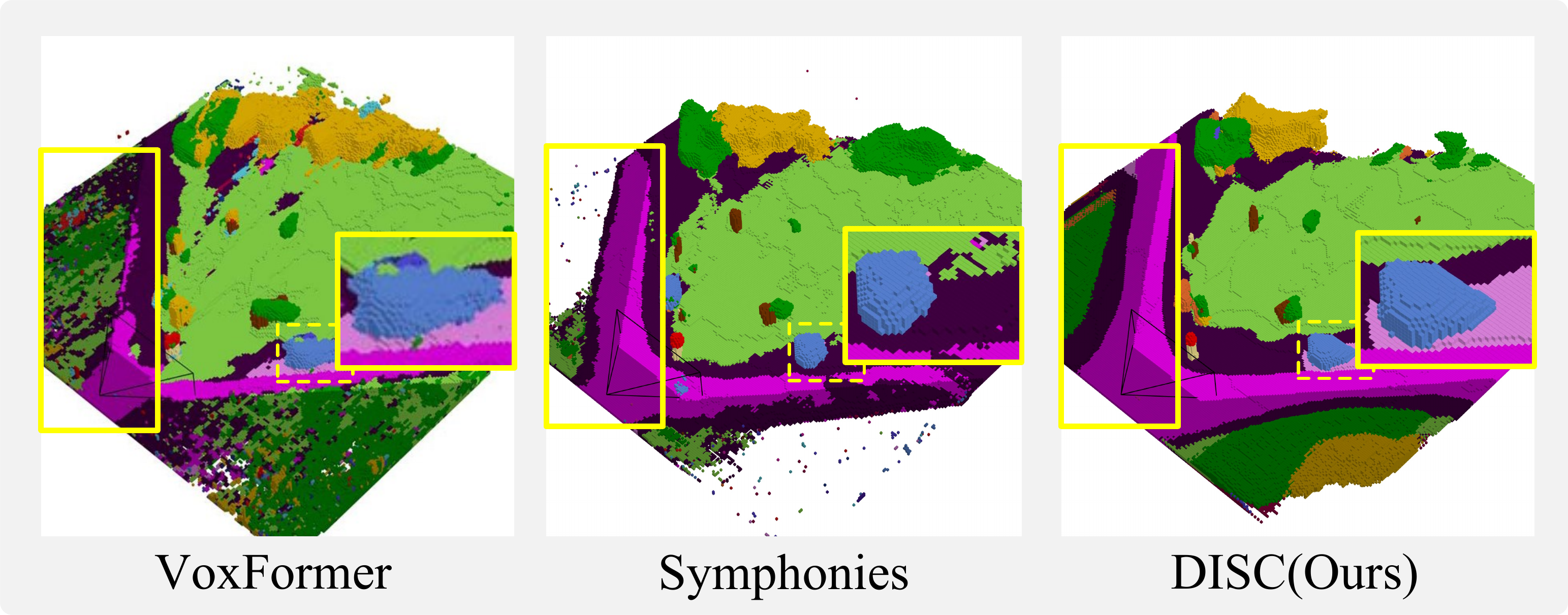}
    \vspace{-0.7cm}
    \caption{\textbf{Comparative Analysis of Different Methods.} Incorporating instance or scene specific information significantly improves prediction accuracy for their corresponding categories.}
    \vspace{-0.35cm}
    \label{fig-2}
\end{figure}

With the rapid development of autonomous driving~\cite{hu2023planning, tian2023occ3d}, accurate environmental perception is crucial for tasks like navigation and obstacle avoidance~\cite{jin2024tod3cap}. The 3D Semantic Scene Completion (SSC) task addresses these critical demands by jointly predicting scene geometry and semantics to enable comprehensive 3D scene understanding. SSC methods are primarily categorized into lidar-based~\cite{roldao2020lmscnet, yan2021sparse, cheng2021s3cnet} and vision-based approaches~\cite{cao2022monoscene, huang2023tri, li2023voxformer, miao2023occdepth, jiang2024symphonize, yu2024context}. Owing to their lower memory consumption, vision-based methods have gained prominence. VoxFormer~\cite{li2023voxformer} pioneers a sparse-to-dense architecture that mitigates projection blurring, and subsequent advances enhance voxel feature learning through techniques like self-distillation~\cite{notallvoxel}, context priors~\cite{yu2024context}, and implicit instance fusion~\cite{jiang2024symphonize}. Since all these methods utilize voxels as the basic units for feature interaction, we classify them as voxel-based methods.


Although voxel-based methods have achieved remarkable performance, we observe distinct prediction flaws in instance and scene categories: instance categories suffer from class omission and semantic ambiguity due to occlusion and projection errors, while scene categories exhibit structural incoherence from out-of-view regions. Based on this analysis, we argue that fully leveraging class-level information can help address the divergent challenges of different categories. As shown in \cref{fig-2}, compared to VoxFormer~\cite{li2023voxformer} (no category-specific priors) and Symphonies~\cite{jiang2024symphonize} (implicit instance priors), our method explicitly integrates both instance and scene information, achieving superior performance in completion details. Nevertheless, despite the proven benefits of class-level information, voxel-based works have not further extended the class-level concepts. We attribute this limitation to three critical challenges faced by voxel-based methods in leveraging class-level information: (1) Voxel construction process inherently disrupts the structural information of categories; (2) Voxel-based methods require a unified module to handle features across all categories, limiting the ability to address divergent challenges posed by instance and scene categories; (3) Voxel-based methods rely on 3D spatial feature interactions, which impose high computational costs and hinder further integration of class-level information.

\begin{figure}[!t]
    \centering
    \includegraphics[width=1.0\linewidth]{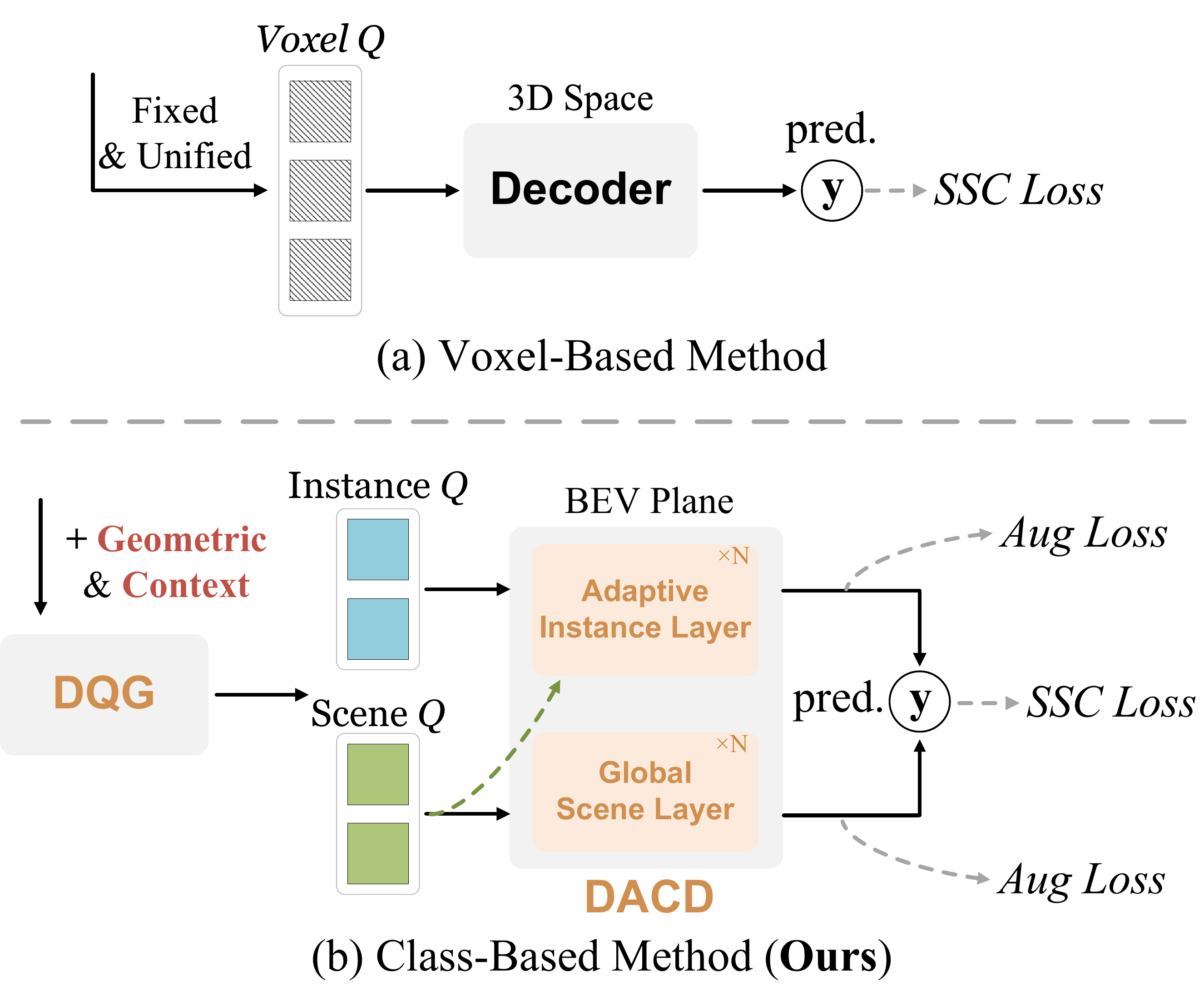}
    \vspace{-0.7cm}
    \caption{\textbf{Comparison of main architectures.} The key differences between our Class-Based Method and the previous Voxel-Based Method are highlighted in \textcolor{brown}{red-brown}. Our method initializes instance and scene queries with semantic and geometric priors, using a dual-stream structure and tailored modules for class-discriminative scene semantic completion.}
    \vspace{-0.35cm}
    \label{fig-1}
\end{figure}

To overcome these challenges, we depart from conventional voxel-based paradigms and propose \textbf{D}isentangling \textbf{I}nstance and \textbf{S}cene \textbf{C}ontexts (\textbf{DISC}), a class-aware dual-stream architecture for discriminative reconstruction. As shown in \cref{fig-1}(b), DISC operates in BEV space and introduces two core modules to leverage category-specific information: a Discriminative Query Generator (DQG) and a Dual-Attention Class Decoder (DACD). Our method addresses three critical issues: (1) Replaces voxel queries with independent instance and scene queries, preserving category-specific structural information while integrating richer geometric and semantic priors; (2) Employs tailored modules to resolve distinct instance and scene challenges; (3) Mitigates BEV’s height-axis limitations by disentangling instance and scene contexts, thereby enabling efficient class-level integration with lower computational costs.

Specifically, our DQG module first initializes independent queries with positional priors based on instance and scene properties, followed by aggregating contextual information to provide semantic priors. Moreover, we argue that feature disentanglement mitigates the BEV limitations in height-axis modeling. For example, road and pedestrian categories within the same BEV grid exhibit significant height disparities, causing vertical feature ambiguity during 3D reconstruction. However, disentangling instance and scene contexts alleviates such multi-height distributions in shared grids. Leveraging this insight, DISC adopts BEV space for query initialization and feature interaction, which significantly reduces memory usage. To address projection errors, we further propose a Coarse-to-fine BEV generation module to deliver precise semantic priors.

Moreover, to design targeted solutions, we further analyze the root causes of instance and scene errors and their relationships. We identify that scene categories suffer from topological errors due to weak global reasoning (e.g., roads appearing in terrain), while instance categories exhibit semantic ambiguities from projection errors and occlusion-induced feature loss. Additionally, scene context aids instance reasoning, for example, pole-shaped objects on sidewalks are more likely to be traffic lights than trunks. Based on these analysis, our DACD decoder implements two specialized layers: the Adaptive Instance Layer (AIL) dynamically fuses image features and scene context to mitigate projection errors and recover occluded details, and the Global Scene Layer (GSL) establishes cross-region interactions to expand the scene's receptive field and ensure layout coherence. Both layers leverage class-specific properties to regulate feature propagation within their respective streams.

Finally, we evaluate DISC on SemanticKITTI and SSCBench-KITTI-360 benchmarks, achieving state-of-the-art (SOTA) mIoU scores of 17.35 and 20.55 respectively. Notably, with only single-frame inputs, DISC outperforms multi-frame SOTA methods. Furthermore, DISC demonstrates substantial gains in instance-level understanding, improving instance mIoU by \textbf{17.9\%} over SOTA methods, while simultaneously achieving SOTA scene mIoU performance on the SemanticKITTI hidden test set.

In summary, our main contributions are the following:
\begin{itemize}
    \item We propose \textbf{DISC}, a novel class-based architecture for the SSC task, featuring a dual-stream structure that decouples instance and scene categories in the BEV plane to fully utilize class-level information.
    \item We propose the Discriminative Query Generator (\textbf{DQG}) that provides geometric and semantic priors for both instance and scene queries, paired with the Dual-Attention Class Decoder (\textbf{DACD}) to address the distinct challenges of instance and scene categories, ensuring accurate and efficient feature interaction within each stream.
    \item We achieve SOTA performance with DISC on both SemanticKITTI and SSCBench-KITTI-360 benchmarks. Moreover, DISC is the first to outperform multi-frame SOTA methods using only single-frame inputs.
\end{itemize}

\section{Related Works}
\label{sec:related_works}

\begin{figure*}[htbp] 
    \centering
    \includegraphics[width=1\textwidth]{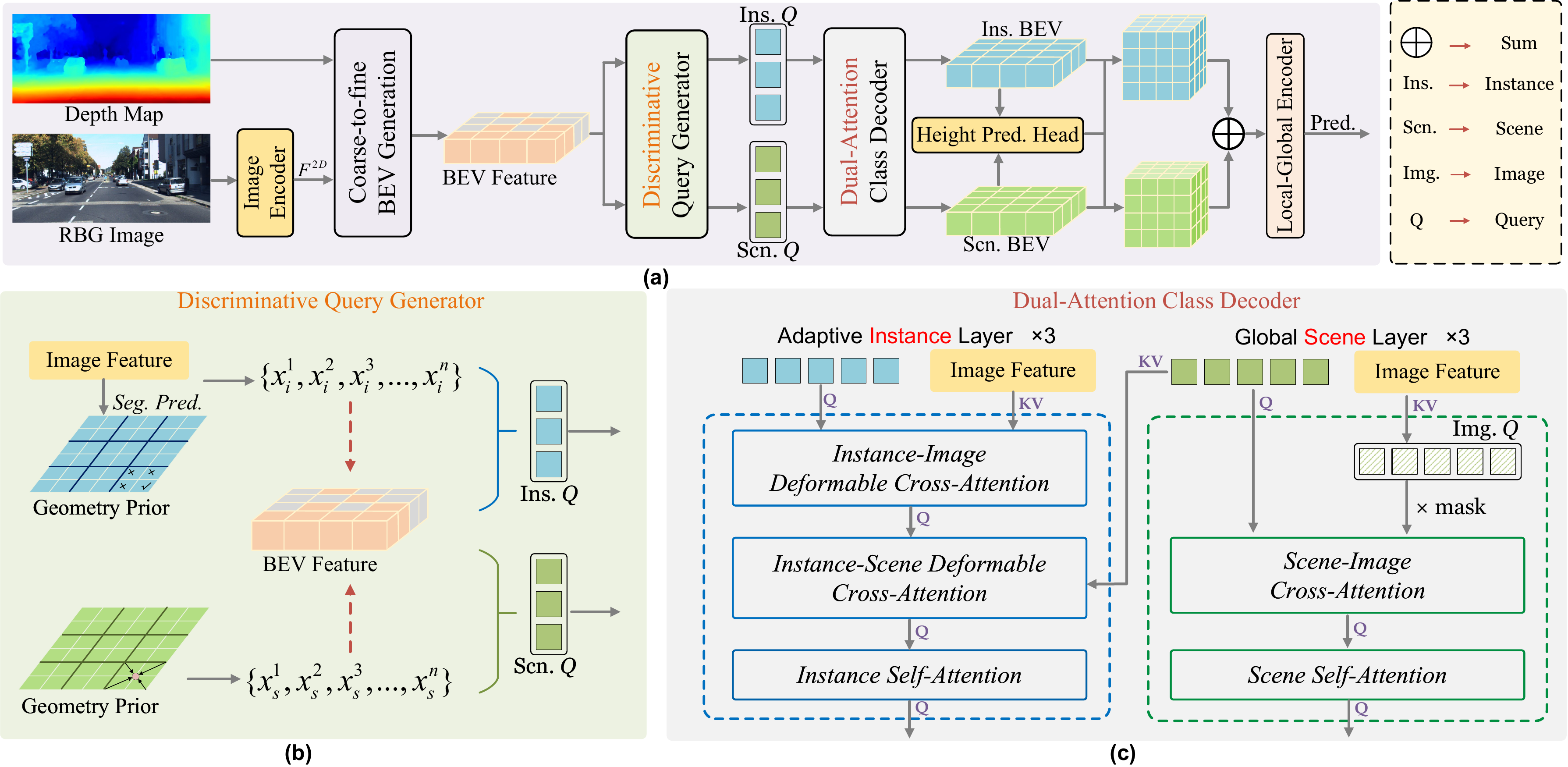}
    \vspace{-0.7cm}
    \caption{\textbf{The overall architecture.} (a) DISC is a novel semantic scene completion method with a dual-stream framework for specialized instance and scene categories processing. (b) The Discriminative Query Generator (DQI) integrates geometric and contextual priors into instance and scene queries based on category attributes. (c) Details of the Adaptive Instance Layer (AIL) and the Global Scene Layer (GSL), which address the distinct challenges faced by instance and scene categories during the reconstruction process in a differentiated manner. For clarity, the Feed-Forward Network (FFN) and positional embedding are omitted in the figure.}
    \label{2_arc}
    \vspace{-0.1cm}
\end{figure*}

\noindent \textbf{3D Semantic Scene Completion.}
The 3D Semantic Scene Completion (SSC) task, introduced by SSCNet~\cite{sscnet}, is an important task in computer vision~\cite{yu2023motrv3, yu2024merlin, li2025ovtr, chen2024delving, chen2025cross} that aims to jointly estimate the complete geometry and semantics of a scene from sparse inputs, playing a crucial role in applications like autonomous driving and virtual reality. Early methods~\cite{roldao2020lmscnet, yan2021sparse, cheng2021s3cnet} focused on LiDAR for 3D semantic occupancy prediction, but camera-based approaches have gained prominence due to their cost efficiency. Recent voxel-based methods, such as MonoScene~\cite{cao2022monoscene}, TPVFormer~\cite{huang2023tri}, and VoxFormer~\cite{li2023voxformer}, enhance semantic information for each voxel, using techniques like monocular image inputs, tri-perspective views, and deformable attention mechanisms. Other approaches like OccDepth~\cite{miao2023occdepth} and VPOcc~\cite{kim2024vpocc} improve geometric projection and address perspective distortion, while Symphonize~\cite{jiang2024symphonize} and CGFormer~\cite{yu2024context} focus on instance-level representations and enhanced voxel features. Voxel-based methods struggle to distinguish instance and scene features, while our approach employs class-aware modeling to resolve this divergence.

\noindent \textbf{BEV-based 3D Perception}
BEV-based methods have gained significant attention in 3D perception due to their compact scene representation~\cite{li2023delving}. These methods can be classified into transformer-based and frustum-based approaches, depending on how BEV maps are derived from camera images. Transformer-based methods~\cite{li2022bevformer, chen2022persformer, harley2023simple, gong2022gitnet, jiang2023polarformer, xu2022cobevt} project BEV grids onto camera images and pull features from neighboring points into the BEV space. Frustum-based methods~\cite{philion2020lift, li2023bevstereo, li2023bevdepth, reading2021categorical, zhang2023sa, wang2024bevspread} lift image features into 3D frustums by predicting depth probabilities and then use voxel pooling to generate BEV features. In this work, we build upon the Lift-Splat-Shoot (LSS) method~\cite{philion2020lift} and leverage depth information to reduce geometric ambiguity caused by depth estimation errors. While BEV cuts computational costs~\cite{hou2024fastocc, yu2023flashocc} for 3D scene completion, it underperforms scene-based methods due to vertical coupling of instance and scene features in unified BEV representations. Our feature-decoupling modules resolve this height-dimension constraint, unlocking BEV's potential.

\section{Method}
\subsection{Overall Architecture}
\label{sec:overview}
The overall architecture of DISC is shown in \cref{2_arc}. A 2D backbone~\cite{jiang2024symphonize} extracts multi-scale feature maps $F^{2D}$, followed by the Discriminative Query Generator (DQG) producing instance queries $\mathbf{Q}_{\text{ins}} \in \mathbb{R}^{N_{\text{ins}}\times C}$ and scene queries $\mathbf{Q}_{\text{scn}} \in \mathbb{R}^{N_{\text{scn}}\times C}$ in the BEV space, where $N_{\text{ins}}$ and $N_{\text{scn}}$ represent the number of initial queries (\cref{sec:DQG}). These queries are processed by the Dual-Attention Class Decoder (DACD) for class-specific feature interaction (\cref{sec:decoder}). Finally, the refined instance and scene features are fused in 3D space and passed through a prediction head to obtain the scene completion results (\cref{sec:fusion&loss}). Detailed descriptions of each component are provided in the following subsections.

\vspace{0.5em} \noindent \textbf{Depth Estimator.} Following previous works~\cite{li2023mask,jiang2024symphonize,yu2024context}, we employ a pretrained MobileStereoNet~\cite{shamsafar2022mobilestereonet} to estimate the depth of the input image. The estimated depths are used to supervise depth predictions and compute the 3D positions of projected points from pixel coordinates.

\subsection{Discriminative Query Generator}
\label{sec:DQG}
As shown in \cref{2_arc}(b), we propose a category-aware query design that distinguishes between instance queries and scene queries to preserve their distinct geometric properties. Unlike uniform voxel queries, our approach integrates geometric and semantic priors into query initialization while maintaining computational efficiency. First, BEV features are derived by projecting 2D features ($F^{2D}$) onto the BEV plane, with projection errors mitigated through a Coarse-to-fine BEV Generation module. For instances, their sparse spatial distribution and projection errors make the image space more suitable for identifying potential positions. However, perspective distortion complicates uniform sampling across different distances. Therefore, we first locate potential instances in the image space and then refine them in the BEV space. For scene queries, we adopt a large-sized design to capture continuous spatial distributions while preserving structural coherence, thereby avoiding the segmentation artifacts caused by fine-grained voxel queries. Notably, all queries incorporate BEV positional embeddings (detailed in the appendix).

\noindent \textbf{Coarse-to-fine BEV Generation. }To address projection errors in LSS~\cite{philion2020lift}, we first generate coarse voxel features $V_{\text{coarse}} \in \mathbb{R}^{C \times X \times Y \times Z}$ via lifting, where $X$, $Y$, and $Z$ correspond to the scene grid dimensions. Depth-guided surface voxels are then refined using $F^{2D}$, producing optimized $V_{\text{fine}} \in \mathbb{R}^{C \times X \times Y \times Z}$ through feature fusion. Finally, BEV features $C \in \mathbb{R}^{C \times X \times Y}$ are obtained by Z-axis max pooling of $V_{\text{fine}}$. Further implementation details are provided in the supplementary materials.

\noindent \textbf{Instance Query.} To improve the recall of instance categories, we detect potential instances through image-space segmentation and project candidate pixels to BEV. A parallelizable neighbor suppression strategy selects $N_{\text{ins}}$ reference positions: $k \times k$ grids form voting blocks, with highest-probability candidates retained per block. This emphasizes small and long-tail objects while maintaining efficiency, expressed as:

\begin{align}
    X_{\text{ins}} &= \big\{\text{CT}(\mathbf{g}_n) \mid \mathbf{g}_n \in \text{Top-}N\big(\{\text{Max}(B_{k\times k}^i) \}_{i=1}^s \big)\big\}
\end{align}
Here, $s$ denotes the number of blocks, and $B_{k \times k}^i$ represents the $i$-th block of $k \times k$ grids, where $i$ ranges from 1 to $s$. The $\text{Max}$ operation selects the grid with the highest probability in each block as a candidate. $\mathbf{g}_n$ refers to the top-N selected candidate grids, and $\text{CT}(\mathbf{g}_n)$ denotes their center coordinates, forming the reference set $X_{\text{ins}}$ for instance queries. The features at these locations in BEV plane initialize the instance queries: $\mathbf{Q}_{\text{ins}} = C[X_{\text{ins}}]$.

\noindent \textbf{Scene Query. } For scene categories,  we design a patch-based scene query generation strategy to enrich the information in each individual query. Specifically, BEV features are divided into multiple equally sized patches, with the center point of each patch serving as the reference point for a scene query. Each patch is upsampled via convolution until the feature size reaches $1\times 1$, which is then used as the initialization feature for the scene query, expressed as:
\begin{align}
    \mathbf{Q}_{\text{scn}} &= \text{UpSample}(C)
\end{align}
Here, $\text{UpSample}$ represents the upsampling operation.

\subsection{Dual-Attention Class Decoder}
\label{sec:decoder}
As discussed in \cref{sec:intro}, existing methods~\cite{li2023voxformer, jiang2024symphonize, yu2024context} employ a unified attention module for both instance and scene features, limiting targeted optimizations. Given the distinct characteristics between instances and scenes, this approach is suboptimal. In contrast, we design task-specific decoding layers tailored to each challenges, as shown in \cref{2_arc} (c). Prone to occlusion and projection errors, instances require multi-level geometric and semantic information for robust reasoning. For example, height direction features and contextual categories facilitate instance refinement and ambiguity reduction. Scene categories, with their continuous distribution, require global receptive fields to capture spatial relationships and maintain layout coherence. Based on these insights, we propose the Dual-Attention Class Decoder comprising an Adaptive Instance Layer and a Global Scene Layer. As follows, we will explain how they achieve targeted feature propagation and interaction.

\begin{figure}[!t]
    \centering
    \includegraphics[width=1.0\linewidth]{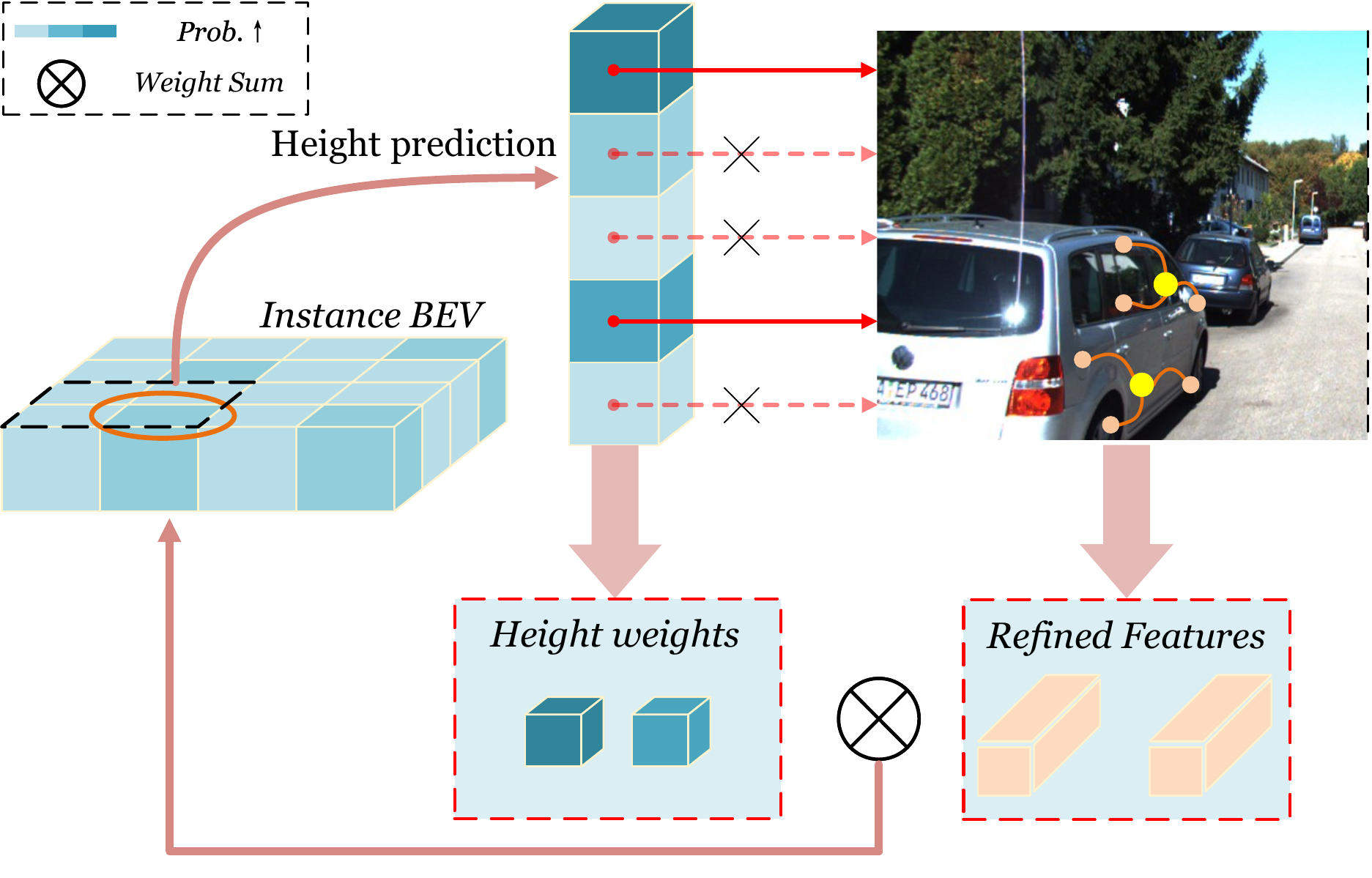}
    \vspace{-0.7cm}
    \caption{\textbf{Instance-Image Cross-Attention.} For each instance query, we adaptively select a series of heights and combine them with its reference point coordinates on the BEV plane to project the query into the image space. This enables capturing image features across multiple height levels.}
    \label{3—height}
\end{figure}

\noindent \textbf{Adaptive Instance Layer.} As shown in \cref{3—height}, to address the loss of height information in BEV space, the instance query is transformed into pillar-like queries. Given the significant height variance within instance classes (e.g., trucks and traffic signs are much taller than pedestrians), we use adaptive height sampling rather than uniform sampling. For each instance query $q_{\text{ins}}$, discrete candidate heights are initialized. A linear layer predicts probabilities for these candidates, and TOP-N selection identifies the most probable heights. These heights are combined with the reference coordinates $x_{\text{ins}}$ to form reference points $P_j = (x_{\text{ins}}, h_j)$. These points are then projected onto the image feature space, where nearby features are sampled using deformable cross attention. The weighted sum of these features is then used to update $q_{\text{ins}}$, expressed as :

\begin{align}
    q_{\text{ins}} = \sum_{j=1}^Nw_j\text{DA}(q_{\text{ins}}, F^{2D}, \mathcal{T}^{WI}(x_{\text{ins}}, h_j))
\end{align}
Here, $h_j$ represents a sampled height, with a total of $N$ heights, and $w_j$ is the corresponding sampling weight. $\mathcal{T}^{WI}$ denotes the projection from the world coordinate system to image space for feature sampling. In practice, the 3D reference points are randomly offset within the voxel grid range to enhance the ability of $q_{\text{ins}}$ to capture features.

After that, the instance queries extract scene features from the region of interest through attention mechanism, which is formulated as: 
\begin{align}
    q_{\text{ins}} = \text{DA}(q_{\text{ins}}, C_{\text{scn}}, x_{\text{ins}})
\end{align}
where $C_{\text{scn}}$ represents scene features output by the GSL. We also utilize self attention to capture internal relationships among instance queries and recover lost information:
\begin{align}
    \mathbf{Q}_{\text{ins}} = \text{SelfAttn}(\mathbf{Q}_{\text{ins}})
\end{align}

Finally, instance features are propagated across the entire BEV plane using a UNet-like~\cite{unet} network, yielding $C_{\text{ins}}$.

\noindent \textbf{Global Scene Layer.} To enable each scene query to capture global information, we construct $\mathbf{Q}_{\text{img}}$ following the same method as $\mathbf{Q}_{\text{scn}}$, obtained by the smallest scale image feature $F^{2D}_s$. For each scene query $q_{\text{scn}}$, global semantic features are aggregated from $\mathbf{Q}_{\text{img}}$ via cross attention, as shown below:
\begin{align}
    q_{\text{scn}} = \text{CrossAttn}(q_{\text{scn}}, \mathbf{Q}_{\text{img}}^{\in\text{Mask}}, \mathbf{Q}_{\text{img}}^{\in\text{Mask}})
\end{align}
where $\text{Mask}$ is a random mask that discards a portion of $\mathbf{Q}_{\text{img}}$, simulating information loss (e.g., occlusions), which aids the network in better inferring the scene layout.

Then $\mathbf{Q}_{\text{scn}}$ interact with BEV features through self attention, further expanding the global receptive field and propagating visible features from nearby areas to distant and unseen regions beyond the viewpoint, as expressed by: 
\begin{align}
    \mathbf{Q}_{\text{scn}} = \text{SelfAttn}(\mathbf{Q}_{\text{scn}})
\end{align}

Finally, the queries are combined into BEV features based on the positions of the reference points, and $C_{\text{scn}}$ is obtained through upsampling layers.

\subsection{Feature Fusion and Losses}
\label{sec:fusion&loss}
\noindent \textbf{Feature Fusion.}
Prior BEV-based 3D reconstruction methods~\cite{yu2023flashocc,hou2024fastocc} suffer from two limitations: (1) oversimplified height features and (2) coupled instance-scene feature interactions. When instance-class objects (e.g., pedestrians) and scene-class elements (e.g., roads) project onto the same BEV grid, their significant height variations create learning conflicts. We address this through category-decoupled height prediction:
\begin{align}
V &= (C_{\text{ins}} \bigotimes H_{\text{ins}}) + (C_{\text{scn}} \bigotimes H_{\text{scn}})
\end{align}
where $H_{\text{ins}}, H_{\text{scn}} \in [0, 1]^{X \times Y \times Z}$  are height distributions predicted via convolutional networks, and $\bigotimes$ denotes broadcasted element-wise multiplication.

We enhance $V$ using a Local and Global Aggregator with Dynamic Fusion to produce the final 3D feature volume. Unlike CGFormer~\cite{yu2024context}, only the left and front views are input to the Global Aggregator, as the BEV plane already contains sufficient detailed features for aggregation.

\vspace{0.5em} \noindent \textbf{Training Loss.}
Following prior works, we use the Scene-Class Affinity Loss $\mathcal{L}_{scal}$ from Monoscene~\cite{cao2022monoscene} for both semantic and geometric predictions, while simultaneously optimizing precision, recall, and specificity. Additionally, a class-frequency weighted cross-entropy loss $\mathcal{L}_{ce}$ is applied, with the total loss for this part given by:
\begin{align}
	\mathcal{L}_{ssc} = \mathcal{L}_{scal}^{geo} + \mathcal{L}^{sem}_{scal} + \mathcal{L}_{ce}
\end{align}
Additionally, we introduce segmentation and height prediction losses for both instance and scene categories on the BEV plane to facilitate discriminative feature learning. Segmentation loss combines cross-entropy and dice loss~\cite{saha2022translating}, while height prediction uses focal loss~\cite{ross2017focal}:
\begin{align}
	\mathcal{L}_{aug} = \mathcal{L}_{seg} + \lambda_h\mathcal{L}_{height} 
\end{align}
The $\mathcal{L}_{aug}$ loss is applied at each decoder layer, halved for layers except the final one. We also use explicit depth loss $\mathcal{L}_{d}$~\cite{li2022bevformer} to supervise the depth prediction. The total loss is:
\begin{align}
    \mathcal{L}_{total} = \lambda _1 \mathcal{L}_{ssc} + \lambda _ 2 \mathcal{L}_{aug} + \lambda_d \mathcal{L}_{d}
\end{align}
In practice,  we set the weights as: $\lambda_1=1$, $\lambda_2=1$, $\lambda_h=5$ and $\lambda_d=0.01$.
\section{Experiments}
\label{sec:experiments}

\begin{table*}[ht]
    \centering
    \newcommand{\clsname}[2]{
        \rotatebox{90}{
            \hspace{-6pt}
            \textcolor{#2}{$\blacksquare$}
            \hspace{-6pt}
            \renewcommand\arraystretch{0.6}
            \begin{tabular}{l}
                #1                                      \\
                \hspace{-4pt} ~\tiny(\semkitfreq{#2}\%) \\
            \end{tabular}
        }}
    \newcommand{\empa}[1]{\textbf{#1}}
    \newcommand{\empb}[1]{\underline{#1}}
    \renewcommand{\tabcolsep}{2pt}
    \renewcommand\arraystretch{1.1}
    \resizebox{\linewidth}{!}
    {
        \begin{tabular}{l|rrr>{\columncolor{gray!20}}r|rrrrrrrrrrrrrrrrrrrr}
            \toprule
            Method                               &
            \multicolumn{1}{c}{IoU}              &
            \multicolumn{1}{c}{InsM}             &
            \multicolumn{1}{c}{ScnM}              &
            mIoU                                 &
            \clsname{road}{road}                 &
            \clsname{sidewalk}{sidewalk}         &
            \clsname{parking}{parking}           &
            \clsname{other-grnd.}{otherground}   &
            \clsname{building}{building}         &
            \clsname{car}{car}                   &
            \clsname{truck}{truck}               &
            \clsname{bicycle}{bicycle}           &
            \clsname{motorcycle}{motorcycle}     &
            \clsname{other-veh.}{othervehicle}   &
            \clsname{vegetation}{vegetation}     &
            \clsname{trunk}{trunk}               &
            \clsname{terrain}{terrain}           &
            \clsname{person}{person}             &
            \clsname{bicyclist}{bicyclist}       &
            \clsname{motorcyclist}{motorcyclist} &
            \clsname{fence}{fence}               &
            \clsname{pole}{pole}                 &
            \clsname{traf.-sign}{trafficsign}
            \\
            \midrule
            \multicolumn{24}{l}{\textit{Temporal Inputs Methods}}                                                                       
            \\
            \hline
        
            VoxFormer-T~\cite{li2023voxformer}       & 43.21  & 4.79 & 22.97 & 13.41 & 54.10 & 26.90 & 25.10 & 7.30 & 23.50 & 21.70 & 3.60 & \empa{1.90} & \empb{1.60} & 4.10 & 24.40 & 8.10 & 24.20 & \empa{1.60} & 1.10 & 0.00 & 13.10 & 6.60 & 5.70 \\
            HASSC-T~\cite{wang2024not}  & 42.87  & 5.27 & 24.51 & 14.38 & 55.30 & 29.60 & 25.90 & 11.30 & 23.10 & 23.00 & 2.90 & \empa{1.90} & 1.50 & 4.90 & 24.80 & 9.80 & \empb{26.50} & \empb{1.40} & \empb{3.00} & 0.00 & 14.30 & 7.00 & 7.10 \\
            H2GFormer-T~\cite{wang2024h2gformer}  & \empb{43.52}  &\empb{5.37} &\empb{25.06} & \empb{14.60} & \empb{57.90} & \empb{30.40} & \empb{30.00} & 6.90 & \empb{24.00} & \empb{23.70} & \empb{5.20} & 0.60 & 1.20 & \empb{5.00} & \empb{25.20} & \empb{10.70} & 25.80 & 1.10 & 0.10 & 0.00 & \empb{14.60} & \empb{7.50} & \textcolor{red}{\empa{9.30}} \\
            HTCL~\cite{HTCL} & \empa{44.23}  & \empa{6.48} & \textcolor{red}{\empa{28.86}} & \empa{17.09} & \textcolor{red}{\empa{64.40}} & \textcolor{red}{\empa{34.80}} & \empa{33.80} & \empa{12.40} & \empa{25.90} & \textcolor{red}{\empa{27.30}} & \empa{5.70} & \empb{1.80} & \empa{2.20} & \empa{5.40} & \empa{25.30} & \empa{10.80} & \textcolor{red}{31.20} & 1.10 & \empa{3.10} & \empa{0.90} & \textcolor{red}{\empa{21.10}} & \textcolor{red}{\empa{9.00}} & \empb{8.30} \\

            \specialrule{0.7pt}{0pt}{0pt}
            \multicolumn{24}{l}{\textit{Single-frame Inputs Methods}}                                                                       
            \\
            \hline
            
            MonoScene$^\ast$~\cite{cao2022monoscene} & 34.16  & 3.59 & 19.4 & 11.08 & 54.70 & 27.10 & 24.80 & 5.70
            & 14.40 & 18.80 & 3.30 & 0.50 & 0.70 & 4.40  & 14.90 & 2.40  & 19.50 & 1.00  & 1.40
            & 0.40  & 11.10 & 3.30 & 2.10         \\
            TPVFormer~\cite{huang2023tri}        &34.25  & 3.49 & 19.88 & 11.26 & 55.10 & 27.20 & 27.40 & 6.50
            & 14.80 & 19.20 & 3.70 & 1.00 & 0.50 & 2.30  & 13.90 & 2.60  & 20.40 & 1.10  & 2.40
            & 0.30  & 11.00 & 2.90 & 1.50 		  \\
            SurroundOcc~\cite{wei2023surroundocc}    & 34.72  & 3.90 & 20.7 & 11.86 & 56.90 & 28.30 & 30.20 & 6.80 
            & 15.20 & 20.60 & 1.40 & 1.60 & 1.20 & 4.40  & 14.90 & 3.40  & 19.30 & 1.40  & 2.00
            & 0.10  & 11.30 & 3.90 & 2.40         \\
            OccFormer~\cite{zhang2023occformer}        & 34.53  & 4.02 & 21.53 & 12.32 & 55.90 & 30.30 & 31.50 & 6.50          
            & 15.70 & 21.60 & 1.20 & 1.50 & 1.70 & 3.20  & 16.80 & 3.90  & 21.30 & 2.20  & 1.10
            & 0.20  & 11.90 & 3.80 & 3.70         \\
            IAMSSC~\cite{xiao2024instance}  & 43.74  & 4.50 & 21.12 & 12.37 & 54.00 & 25.50 & 24.70 & 6.90 & 19.20 & 21.30 & 3.80 & 1.10 & 0.60 & 3.90 & 22.70 & 5.80 & 19.40 & 1.50 & 2.90 & 0.50 & 11.90 & 5.30 & 4.10 \\
            VoxFormer-S~\cite{li2023voxformer}      & 42.95  & 4.39 & 20.89 & 12.20 & 53.90 & 25.30 & 21.10 & 5.60
            & 19.80 & 20.80 & 3.50 & 1.00 & 0.70 & 3.70  & 22.40 & 7.50  & 21.30 & 1.40  & 2.60 
            & 0.20  & 11.10 & 5.10 & 4.90         \\
            DepthSSC~\cite{yao2023depthssc}          & \empb{44.58}  & 4.87 & 22.26 & 13.11 & 55.64 & 27.25 & 25.72 & 5.78
            & 20.46 & 21.94 & 3.74 & 1.35 & 0.98 & 4.17  & 23.37 & 7.64  & 21.56 & 1.34  & 2.79
            & 0.28  & 12.94 & 5.87 & 6.23         \\
            Symphonize~\cite{jiang2024symphonize}	  & 42.19  & 6.14 & 24.93 & 15.04 & 58.40 & 29.30 & 26.90 & 11.70
            & 24.70 & 23.60 & 3.20 & 3.60 & 2.60 & 5.60  & 24.20 & 10.00 & 23.10 & \textcolor{red}{\empa{3.20}}  & 1.90  
            & \textcolor{red}{\empa{2.00}}  & 16.10 & 7.70 & 8.00         \\
            HASSC-S~\cite{wang2024not} & 43.40  & 5.12 & 22.46 & 13.34 & 54.60 & 27.70 & 23.80 & 6.20 & 21.10 & 22.80 & 4.70 & 1.60 & 1.00 & 3.90 & 23.80 & 8.50 & 23.30 & 1.60 & \empb{4.00} & 0.30 & 13.10 & 5.80 & 5.50 \\
            StereoScene~\cite{li2023stereoscene}  & 43.34  & 5.73 & 26.04 & 15.36 & 61.90 & 31.20 & 30.70 & 10.70 & 24.20 & 22.80 & 2.80 &  3.40 & 2.40 & \empb{6.10} & 23.80 & 8.40 & 27.00 & \empb{2.90} & 2.20 & 0.50 & 16.50 & 7.00 & 7.20 \\
            H2GFormer-S~\cite{wang2024h2gformer} & 44.20  & 5.05 & 23.33 & 13.72 & 56.40 & 28.60 & 26.50 & 4.90 & 22.80 & 23.40 & 4.80 & 0.80 & 0.90 & 4.10 & 24.60 & 9.10 & 23.80 & 1.20 & 2.50 & 0.10 & 13.30 & 6.40 & 6.30 \\
            MonoOcc-S~\cite{zheng2024monoocc}  & -  & 5.36 & 23.14 & 13.80 & 55.20 & 27.80 & 25.10 & 9.70 & 21.40 & 23.20 & 5.20 & 2.20 & 1.50 & 5.40 & 24.00 & 8.70 & 23.00 & 1.70 & 2.00 & 0.20 & 13.40 & 5.80 & 6.40 \\
            MonoOcc-L~\cite{zheng2024monoocc} & -  &6.95 & 25.29 & 15.63 & 59.10 & 30.90 & 27.10 & 9.80 & 22.90 & 23.90 & \textcolor{red}{\empa{7.20}} & \textcolor{red}{\empa{4.50}} & 2.40 & 7.70 & 25.00 & 9.80 & \empb{26.10} & 2.80 & \textcolor{red}{\empa{4.70}} & 0.60 & 16.90 & 7.30 & 8.40 \\
            VPOcc~\cite{kim2024vpocc} & 44.09  &6.03 &26.31 & 15.65 & 59.10 & 32.30 & 30.90 & 9.70 & \empb{26.30} & 24.40 & 5.30 & 3.30 & \empb{3.20} & 5.60 & \empb{25.90} & 9.70 & 25.70 & 2.40 & 2.90 & 0.30 & 17.20 & 6.60 & 6.30 \\
            CGFormer~\cite{yu2024context} & 44.41  &\empb{6.15} &\empb{28.24} & \empb{16.63} & \empa{64.30} & \empb{34.20} & \empb{34.10} & \empb{12.10} & 25.80 & \empb{26.10} & 4.30 & 3.70 & 1.30 & 2.70 & 24.50 & \textcolor{red}{\empa{11.20}} & \empa{29.30} & 1.70 & 3.60 & 0.40 & \empb{18.70} & \empa{8.70} & 
            \textcolor{red}{\empa{9.30}} \\
            \hline
            \empa{DISC(ours)} & \textcolor{red}{\empa{45.32}} & \textcolor{red}{\empa{7.25}} & \empa{28.56} & \textcolor{red}{\empa{17.35}}   & \empb{63.10} & \empa{34.70} & \textcolor{red}{\empa{34.60}} & \textcolor{red}{\empa{12.60}} & \textcolor{red}{\empa{26.60}} &
            \empa{26.70} & \empb{5.50} & \empb{4.00} & \textcolor{red}{\empa{4.70}} & \textcolor{red}{\empa{8.10}} & \textcolor{red}{\empa{26.50}} & \empb{10.30} & \empa{29.30} & 2.80 & 2.50 & \empb{1.10} & \empa{19.30} & \empb{8.40} & \empb{8.70}\\
            \bottomrule
        \end{tabular}
    }
    \vspace{-0.1cm}
    \caption{\textbf{Quantitative results on SemanticKITTI \texttt{test}.} $^\ast$ represents the reproduced results from~\cite{huang2023tri,zhang2023occformer}. Among all methods, the top three ranked approaches are marked as \textcolor{red}{\empa{red}}, \empa{bold}, and \empb{underlined}. For single-frame methods, DISC achieves SOTA performance in mIoU, IoU, InsM, and ScnM. Notably, using only single-frame input, DISC surpasses even multi-frame SOTA methods in mIoU, IoU, and InsM.}
    \label{tab:sem_kitti_test}
    \vspace{-0.1cm}
\end{table*}


\subsection{Dataset and Metric}
DISC is evaluated on two widely used datasets: SemanticKITTI~\cite{behley2019semantickitti} and SSCBench-KITTI-360~\cite{li2023sscbench}. For evaluation metrics, we employ standard measures such as Intersection over Union (IoU) and mean IoU (mIoU) for voxel-wise predictions. Additionally, we introduce instance mean IoU (InsM) and scene mean IoU (ScnM) metrics to assess the model's perceptual capabilities across different categories. Further details regarding the datasets and metrics are provided in the supplementary materials.

\subsection{Implementation Details}
Following Symphonies~\cite{jiang2024symphonize}, the ResNet-50~\cite{he2016deep} backbone and image encoder are initialized with pre-trained MaskDINO~\cite{li2023mask} weights. Generally, categories such as car, bicycle, and traffic sign are classified as instances, while road, sidewalk, and building are categorized as scene. Detailed category definitions are provided in the supplementary materials. In our experiments, we observed that our network converges faster compared to previous works, allowing us to reduce the total training epochs to 20 which is shorter than most existing approaches. We use the AdamW~\cite{loshchilov2017decoupled} optimizer with an initial learning rate of 2e-4 and a weight decay of 1e-4. The learning rate is reduced by a factor of 0.1 at the 12th epoch.

\subsection{Comparison with state-of-the-art}  
\cref{tab:sem_kitti_test} presents our results on the SemanticKITTI hidden test set. DISC outperforms all competing methods in IoU, mIoU, InsM, and ScnM, achieving scores of 45.30, 17.35, 7.25, and 28.56, respectively. Notably, DISC shows significant progress in instance categories, surpassing the existing state-of-the-art (SOTA) method by 17.9\%, strongly validating the effectiveness of our instance-specific design. Additionally, DISC is the first single-frame-based method to outperform SOTA multi-frame fusion methods in IoU, mIoU, and InsM, while achieving a ScnM score only 0.30 lower, demonstrating its ability to fully exploit single-frame information. DISC achieves the best performance in most classes, such as sidewalk, parking, building, motorcycle, and other-vehicle.

We also conducted experiments on SSCBench-KITTI-360. As shown in \cref{tab:kitti_360_test}, DISC achieves excellent performance with an mIoU of 20.55, IoU of 47.35, InsM of 13.47, and ScnM of 28.88. Moreover, DISC outperforms all camera-based methods and LiDAR-based methods in mIoU and InsM. This analysis further confirms the effectiveness and outstanding performance of DISC.
\subsection{Ablation Studies}
\begin{table*}[ht]
    \centering
    \newcommand{\clsname}[2]{
        \rotatebox{90}{
            \hspace{-6pt}
            \textcolor{#2}{$\blacksquare$}
            \hspace{-6pt}
            \renewcommand\arraystretch{0.6}
            \begin{tabular}{l}
                #1                                       \\
                \hspace{-4pt} ~\tiny(\sscbkitfreq{#2}\%) \\
            \end{tabular}
        }}
    \newcommand{\empa}[1]{\textbf{#1}}
    \newcommand{\empb}[1]{\underline{#1}}
    \renewcommand{\tabcolsep}{2pt}
    \renewcommand\arraystretch{1.2}
    \scalebox{0.78}
    {
        \begin{tabular}{l|rrr>{\columncolor{gray!20}}r|rrrrrrrrrrrrrrrrrr}
            \toprule
            \multicolumn{1}{c|}{Method}                                 &
            \multicolumn{1}{c}{IoU}                                     &
            \multicolumn{1}{c}{InsM}                                    &
            \multicolumn{1}{c}{ScnM}                                   &
            mIoU                                                         &
            \multicolumn{1}{c}{\clsname{car}{car}}                      &
            \multicolumn{1}{c}{\clsname{bicycle}{bicycle}}              &
            \multicolumn{1}{c}{\clsname{motorcycle}{motorcycle}}        &
            \multicolumn{1}{c}{\clsname{truck}{truck}}                  &
            \multicolumn{1}{c}{\clsname{other-veh.}{othervehicle}}      &
            \multicolumn{1}{c}{\clsname{person}{person}}                &
            \multicolumn{1}{c}{\clsname{road}{road}}                    &
            \multicolumn{1}{c}{\clsname{parking}{parking}}              &
            \multicolumn{1}{c}{\clsname{sidewalk}{sidewalk}}            &
            \multicolumn{1}{c}{\clsname{other-grnd.}{otherground}}      &
            \multicolumn{1}{c}{\clsname{building}{building}}            &
            \multicolumn{1}{c}{\clsname{fence}{fence}}                  &
            \multicolumn{1}{c}{\clsname{vegetation}{vegetation}}        &
            \multicolumn{1}{c}{\clsname{terrain}{terrain}}              &
            \multicolumn{1}{c}{\clsname{pole}{pole}}                    &
            \multicolumn{1}{c}{\clsname{traf.-sign}{trafficsign}}       &
            \multicolumn{1}{c}{\clsname{other-struct.}{otherstructure}} &
            \multicolumn{1}{c}{\clsname{other-obj.}{otherobject}}
            \\
            \midrule
            \multicolumn{23}{l}{\textit{LiDAR-based methods}}                                                                                                                                                                                                                                                                                                                               \\
            \hline
            SSCNet~\cite{sscnet}        & \textcolor{red}{\empa{53.58}}        & 4.96        & \textcolor{red}{\empa{28.88}}      & 16.95        
            & \textcolor{red}{\empa{31.95}}      & 0.00      & 0.17      & 10.29     & 0.00      & 0.07      & \textcolor{red}{\empa{65.70}} 
            & \empa{17.33}      & \textcolor{red}{\empa{41.24}}     & 3.22      & \textcolor{red}{\empa{44.41}}     & 6.77      & \textcolor{red}{\empa{43.72}}     & \textcolor{red}{\empa{28.87}} 
            & 0.78       & 0.75      & 8.69      & 0.67         \\
            LMSCNet~\cite{roldao2020lmscnet}      & \empb{47.35}     & 2.42      & 24.96      & 13.65         
            & 20.91     & 0.00      & 0.00      & 0.26      & 0.58      & 0.00      & \empb{62.95}        
            & 13.51     & 33.51     & 0.20      & \empa{43.67}     & 0.33      & \empa{40.01}     & \empa{26.80}        
            & 0.00      & 0.00      & 3.63      & 0.00         \\
            \specialrule{0.7pt}{0pt}{0pt}
            \multicolumn{23}{l}{\textit{Camera-based methods}}                                                                                                                                                                                                                                                                                                                              \\
            \hline
            MonoScene~\cite{cao2022monoscene}      & 37.87        & 5.22       & 19.41       & 12.31        
            & 19.34        & 0.43        & 0.58        & 8.02         & 2.03         & 0.86        
            & 48.35        & 11.38        & 28.13        & 3.32        & 32.89        & 3.53        
            & 26.15        & 16.75        & 6.92         & 5.67        & 4.20         & 3.09         \\
            TPVFormer~\cite{huang2023tri}      & 40.22        & 5.89      & 21.39      & 13.64         
            & 21.56        & 1.09        & 1.37        & 8.06         & 2.57         & 2.38        
            & 52.99        & 11.99        & 31.07        & 3.78        & 34.83        & 4.80        
            & 30.08        & 17.52        & 7.46         & 5.86        & 5.48         & 2.70         \\
            VoxFormer~\cite{li2023voxformer}      & 38.76        & 4.89       & 18.93        & 11.81         
            & 17.84        & 1.16        & 0.89        & 4.56         & 2.06         & 1.63        
            & 47.01        & 9.67         & 27.21        & 2.89        & 31.18        & 4.97        
            & 28.99        & 14.69        & 6.51         & 6.92        & 3.79         & 2.43         \\
            OccFormer~\cite{zhang2023occformer}      & 40.27        & 6.76      & 22.39        & 13.81        
            & 22.58        & 0.66        & 0.26        & 9.89         & 3.82         & 2.77       
            & 54.30        & 13.44        & 31.53        & 3.55        & 36.42       & 4.80        
            & 31.00        & 19.51        & 7.77         & 8.51        & 6.95         & 4.60         \\
            DepthSSC~\cite{yao2023depthssc}     & 40.85         & 7.19    & 21.72      & 14.28       
            & 21.90        & \empb{2.36}         & \empb{4.30}        & 11.51         & 4.56      & 2.92 
            & 50.88        & 12.89        & 30.27       & 2.49          & 37.33     & 5.22 
            & 29.61        & 21.59        & 5.97        & 7.71          & 5.24      & 3.51           \\
            Symphonies~\cite{jiang2024symphonize}       & 44.12     & \empa{13.11}     & 24.05   & \empb{18.58} 
                & \empa{30.02}        & 1.85         & \empa{5.90}        & \textcolor{red}{\empa{25.07}}        & \textcolor{red}{\empa{12.06}}     & \textcolor{red}{\empa{8.20}} 
            & 54.94        & 13.83        & 32.76       & \textcolor{red}{\empa{6.93}}          & 35.11     & \empa{8.58} 
                & 38.33        & 11.52        & \empb{14.01}       & \empb{9.57}          & \textcolor{red}{\empa{14.44}}     & \textcolor{red}{\empa{11.28}}          \\
            CGFormer~\cite{yu2024context}       & \empa{48.07}     & \empb{12.08}    & \empa{28.01}      & \empa{20.05} 
            & \empb{29.85}        & \empa{3.42}         & 3.96        & \empb{17.59}         & \empb{6.79}      & \empb{6.63} 
            & \empa{63.85}        & \empb{17.15}        & \empa{40.72}       & \empa{5.53}          & \empb{42.73}     & \empb{8.22} 
            & \empb{38.80}        & \empb{24.94}        & \empa{16.24}       & \empa{17.45}         & \empb{10.18}     & \empb{6.77}           \\

            \hline
            \empa{DISC(ours)}   & \empb{47.35}    &\textcolor{red}{\empa{13.47}}      & \empb{27.63}         &  \textcolor{red}{\empa{20.55}}
            & 29.41        & \textcolor{red}{\empa{4.64}}       & \textcolor{red}{\empa{8.27}}      &\empa{19.24}    &\empa{8.51}       &\empa{6.74}
            & 61.88        &\textcolor{red}{\empa{17.56}}       & \empb{40.09}     & \empb{5.27}      &42.53      &\textcolor{red}{\empa{9.24}}
            & 38.76        &23.05       & \textcolor{red}{\empa{16.73}}     & \textcolor{red}{\empa{19.51}}     &\empa{10.32}      & \empa{8.21}\\
            \bottomrule
        \end{tabular}
    }
    \vspace{-0.1cm}
        \caption{\textbf{Quantitative results on SSCBench-KITTI360 \texttt{test}.} The results for most counterparts are provided in \cite{li2023sscbench}. Among all methods, the top three ranked approaches are marked as \textcolor{red}{\empa{red}}, \empa{bold}, and \empb{underlined}. DISC achieves SOTA results in mIoU and InsM, while surpassing LiDAR-based methods across multiple category-specific metrics.}
    \label{tab:kitti_360_test}
\end{table*}

\begin{table}[ht]
    \centering
    \newcommand{\gray}[1]{\textcolor{gray}{#1}} 
    {
        \begin{tabular}{l|ccc>{\columncolor{gray!20}}c}
            \toprule

            Method                 & \multicolumn{1}{c}{IoU$\uparrow$}        & \multicolumn{1}{c}{InsM$\uparrow$} & \multicolumn{1}{c}{ScnM$\uparrow$} & \multicolumn{1}{>{\columncolor{gray!20}}c}{mIoU$\uparrow$} \\
            \midrule
            Baseline             & 20.05       & 6.06      & 19.14     & 12.69 \\
            + Instance Stream  & 45.12       & \textbf{9.79}      & 24.73   & 16.86 \\
            + Scene Stream & 45.85             & 8.55      & 25.80    & 16.73 \\
            \hline
            DISC(Ours)   & \textbf{45.93}      & 8.75      & \textbf{26.27}   & \textbf{17.05}\\
            \bottomrule
        \end{tabular}
    }
    \vspace{-0.1cm}
    \caption{\textbf{Ablation study on architecture of DISC.} DISC achieves optimal comprehensive performance.}
    \label{tab:ablat_arcs}
\end{table}
\begin{table}[ht]
    \centering
    \newcommand{\gray}[1]{\textcolor{gray}{#1}} 
    {
        \begin{tabular}{c|cc>{\columncolor{gray!20}}c}
            \toprule

            Method    & \multicolumn{1}{c}{InsM$\uparrow$} & \multicolumn{1}{c}{ScnM$\uparrow$} & \multicolumn{1}{>{\columncolor{gray!20}}c}{mIoU$\uparrow$}\\
            \midrule
            w/o C2FBEV              &  8.56   &   25.74  &  16.70\\
            w/o Image-assisted strategy     & 8.44  &  25.75  & 16.64 \\
            Using a fixed threshold        & 8.67      & 26.00  & 16.88 \\
            \hline
            DISC instance query    & \textbf{8.75}  & \textbf{26.27} & \textbf{17.05} \\
            \bottomrule
        \end{tabular}
    }
    \vspace{-0.1cm}
    \caption{\textbf{Ablation study on query generator.} We evaluate diverse query generation strategies for optimal performance.}
    \label{tab:abla_insq}
    \vspace{-0.3cm}
\end{table}
\noindent \textbf{Analysis of architecture. }\cref{tab:ablat_arcs} analyzes different architectures. The baseline can be regarded as a simplified FlashOcc~\cite{yu2023flashocc}, which consists of a 2D backbone for extracting image features, a Coarse-to-fine BEV generation module for producing reliable BEV features, a BEV encoder for further processing, and a 3D global and local encoder for handling 3D features, similar to CGFormer~\cite{yu2024context}. On top of the baseline, adding the instance stream and scene stream individually leads to significant improvements of 3.73 InsM and 6.66 ScnM for instance and scene, respectively. When integrating both the instance and scene streams, DISC achieves a balanced improvement in instance and scene metrics and delivers optimal IoU and mIoU scores, further validating the effectiveness of our methods.



\begin{table}[ht]
        \centering
        \begin{tabular}{ccc|c>{\columncolor{gray!20}}c}
            \toprule
            Ins. SA & Ins.-Img. CA & Ins.-Scn. CA & InsM  & mIoU           \\
            \midrule
            \texttimes      &   \texttimes     &\texttimes      &  8.55        & 16.73 \\
            \checkmark      &   \texttimes     &     \texttimes &   8.62       &   16.69        \\
            \checkmark      & \checkmark   &   \texttimes  &      8.73     &   16.98        \\
            \checkmark      &    \texttimes   & \checkmark   &    8.69      &   16.86        \\
            \hline
            \checkmark      & \checkmark    & \checkmark & \textbf{8.75} & \textbf{17.05} \\
            \bottomrule
        \end{tabular}
        \vspace{-0.1cm}
        \caption{\textbf{Ablation study on the Adaptive Instance Layer.} Cross-attention with image plays a crucial role in instance performance.}
        \label{tab:ins_decoder}
\end{table}
            
\begin{table}[ht]
        \centering
        \begin{tabular}{cc|c>{\columncolor{gray!20}}c}
            \toprule
            Scn. SA &  Scn.-Img. CA & ScnM  & mIoU           \\
            \midrule
            \texttimes       &    \texttimes   &     24.73       & 16.86   \\
            \checkmark      &   \texttimes     &    25.99       &   17.01        \\
            \texttimes      & \checkmark  &   25.96        & 16.85      \\
            \hline
            \checkmark      & \checkmark   & \textbf{26.27}          & \textbf{17.05}      \\
            \bottomrule
        \end{tabular}
        \vspace{-0.1cm}
        \caption{\textbf{Ablation study on the Global Scene Layer.} Self-attention proves critical for scene reconstruction.}
        \vspace{-0.3cm}
        \label{tab:bk_decoder}
\end{table}
\noindent \textbf{Analysis of discriminative query generator.}
\cref{tab:abla_insq} provide an ablation analysis of the Discriminative Query Generator (DQG). \cref{tab:abla_insq} shows that removing the Coarse-to-fine BEV generation module significantly reduces both InsM and ScnM. Additionally, eliminating the front-view-assisted candidate point localization strategy and replacing the grid-based candidate point selection strategy with a fixed threshold has a more pronounced impact on instance than on scene. Furthermore, we show that a patch size of 4 for scene query initialization yields the best mIoU and ScnM (see the appendix for details). These results highlight the effectiveness of the DQG in providing sufficient geometric and semantic priors for query initialization, leading to optimal network performance.

\begin{figure*}[htbp] 
    \centering
    \includegraphics[width=1\textwidth]{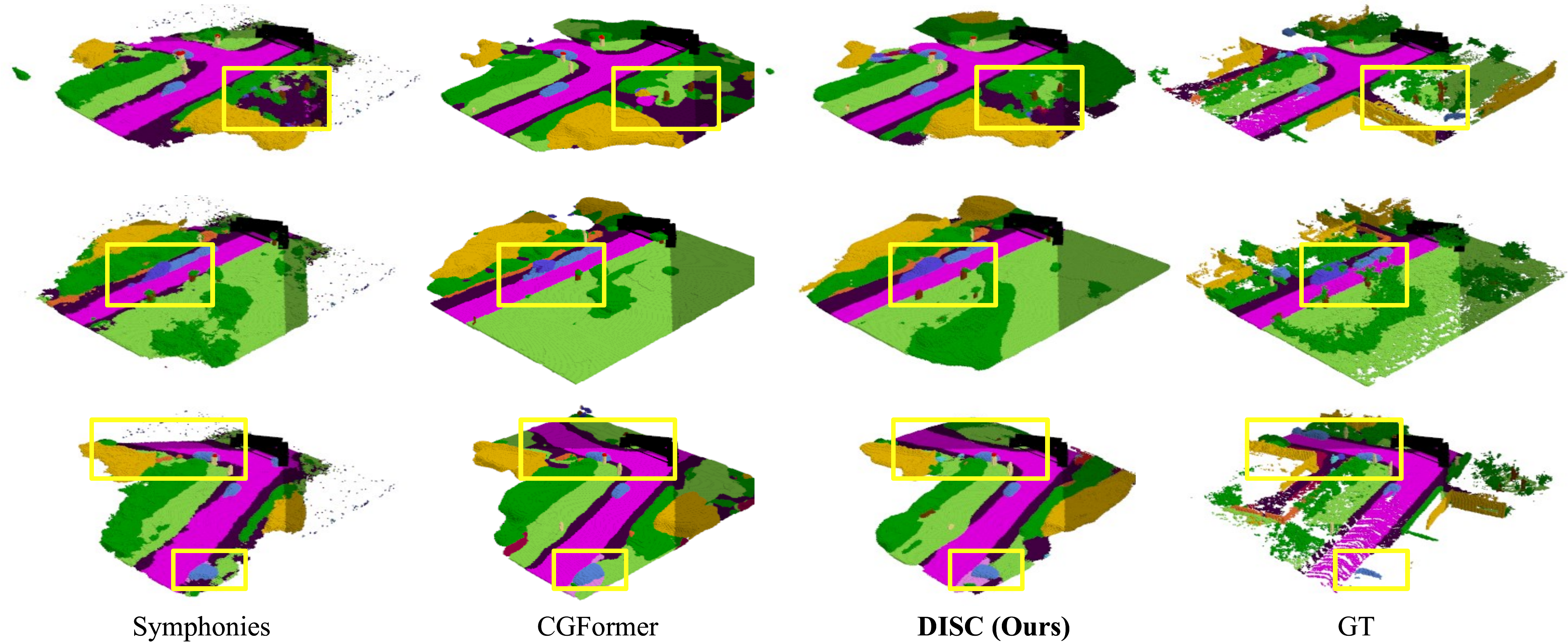}
    \caption{\textbf{Qualitative visualization results on the SemanticKITTI~\cite{behley2019semantickitti} validation set.} Compared to state-of-the-art (SOTA) methods, DISC produces more logical scene layouts and more accurate and detailed instance predictions.}
    \label{6-vis1}
\end{figure*}
\noindent \textbf{Analysis of dual-attention class decoder. }
To analyze the flow of instance and scene queries within their respective decoder layers, we evaluate the interactions between various modules in Adaptive Instance Layer (AIL) and the Global Scene Layer (GSL). \cref{tab:ins_decoder} shows that removing the instance-image cross-attention has the most significant impact on performance, underscoring the importance of fusing instance queries with front-view features. \cref{tab:bk_decoder} reveals that scene categories rely more on the self attention module to refine the global scene layout compared to instance categories, highlighting the distinct reconstruction challenges faced by each category.

\subsection{Visualizations}
\noindent \textbf{Qualitative Results. }
\cref{6-vis1} visualizes predicted results on the SemanticKITTI validation set from Symphonies~\cite{jiang2024symphonize}, CGFormer~\cite{yu2024context}, and our proposed DISC. We compare the performance differences between DISC and other SOTA methods on scene and instance categories. As highlighted by the yellow boxes, DISC predicts more logical scene layouts (e.g., roads do not intersect with terrain) and better geometric distributions (e.g., road structures are more complete and realistic), leveraging the scene features' global perception capabilities. For instances, DISC mitigates the impact of projection errors and occlusions, resulting in more accurate and detailed instance predictions. Additional qualitative results can be found in the Appendix.

\begin{figure}[!t]
    \centering
    \includegraphics[width=1.0\linewidth]{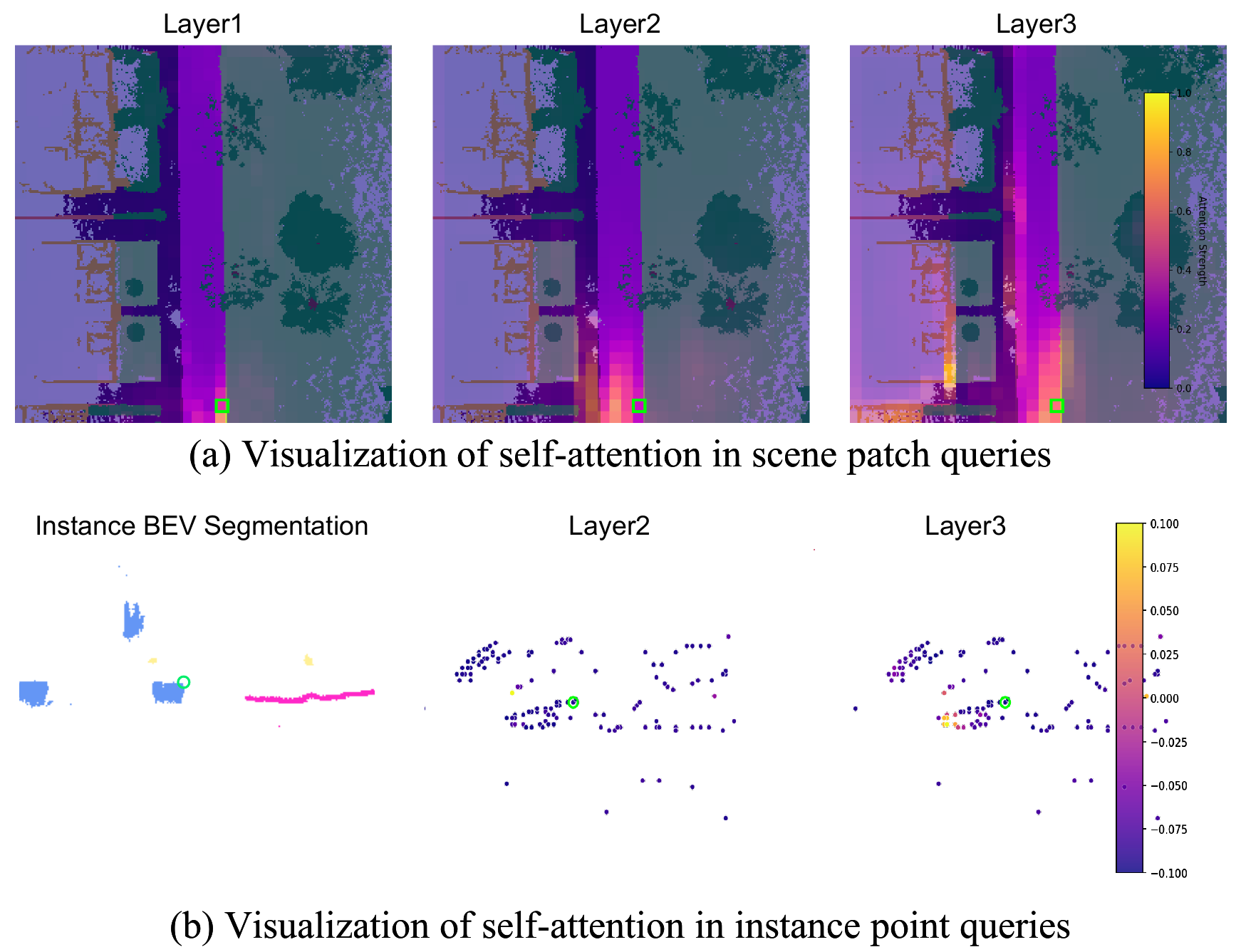}
    \vspace{-0.7cm}
    \caption{\textbf{Analysis of attention maps within DISC. }}
    \vspace{-0.3cm}
    \label{6-attnvis}
\end{figure}

\noindent \textbf{Attention Map Analysis. }In \cref{6-attnvis}, we visualize the interaction mechanisms of scene and instance queries within their respective layers. The patch-based scene queries expand their receptive field through multi-layer self-attention. \cref{6-attnvis} (b) shows the ground truth distribution of instance categories and the instance query candidate points generated by the Discriminative Query Generator, which are concentrated around the ground truth positions. Additionally, in instance self-attention, queries of the same category receive higher attention weights. These results highlight the effectiveness of our discriminative processing for scene and instance categories.

\section{Conclusion}
In this paper, we propose \textbf{DISC}, a dual-stream neural architecture that resolves semantic completion challenges through disentangled refinement of instance and scene representations on the BEV plane. The framework incorporates an efficient query generator (\textbf{DQG}) that fuses geometric-semantic features to enhance both instance and scene queries, complemented by a dual-attention class decoder (\textbf{DACD}) comprising a global-aware scene layer for contextual ambiguity reduction and an adaptive instance layer employing dynamic feature aggregation to address occlusion and projection errors. Experimental validation demonstrates that DISC achieves \textbf{state-of-the-art} performance on SemanticKITTI and SSCBench-KITTI-360 benchmarks while maintaining computational efficiency.

\section*{Acknowledgments}
This work was supported by the National Natural Science Foundation of China under Grant 62176096.
{
    \small
    \bibliographystyle{ieeenat_fullname}
    \bibliography{main}
}


\end{document}